\documentclass[conference]{IEEEtran}
\IEEEoverridecommandlockouts
\usepackage{cite}
\usepackage{amsmath,amssymb,amsfonts}
\usepackage{graphicx}
\usepackage{textcomp}
\usepackage{xcolor}

\usepackage{algorithm,algpseudocode}
\usepackage{hyperref}
\def\BibTeX{{\rm B\kern-.05em{\sc i\kern-.025em b}\kern-.08em
    T\kern-.1667em\lower.7ex\hbox{E}\kern-.125emX}}
\begin{document}

\title{Example Mining for Incremental Learning in Medical Imaging\\
\thanks{*paralleldots.xyz}
}

\author{\IEEEauthorblockN{Pratyush Kumar}
\IEEEauthorblockA{\textit{}
\textit{ParallelDots, Inc.*}\\
Gurugram, India \\
pratyush@paralleldots.com}
\and
\IEEEauthorblockN{Muktabh Mayank Srivastava}
\IEEEauthorblockA{\textit{}
\textit{ParallelDots, Inc.*}\\
Gurugram, India \\
muktabh@paralleldots.com}
}

\maketitle

\begin{abstract}
Incremental Learning is well known machine learning approach wherein the weights of the learned model are dynamically and gradually updated to generalize on new unseen data without forgetting the existing knowledge. Incremental learning proves to be time as well as resource-efficient solution for deployment of deep learning algorithms in real world as the model can automatically and dynamically adapt to new data as and when annotated data becomes available. The development and deployment of Computer Aided Diagnosis (CAD) tools in medical domain is another scenario, where incremental learning becomes very crucial as collection and annotation of a comprehensive dataset spanning over multiple pathologies and imaging machines might take years. However, not much has so far been explored in this direction. In the current work, we propose a robust and efficient method for incremental learning in medical imaging domain. Our approach makes use of Hard Example Mining technique (which is commonly used as a solution to heavy class imbalance) to automatically select a subset of dataset to fine-tune the existing network weights such that it adapts to new data while retaining existing knowledge. We develop our approach for incremental learning of our already under test model for detecting dental caries. Further, we apply our approach to one publicly available dataset and demonstrate that our approach reaches the accuracy of training on entire dataset at once, while availing the benefits of incremental learning scenario.\\
\end{abstract}


\section{Introduction}

Recent years have seen an uprising trend in the use of Convolution Neural Networks (CNNs) based algorithms being used in computer vision tasks such as image classification, object detection, and semantic segmentation. There also exist compelling evidences for remarkable performance of CNNs in medical image segmentation and classification tasks. These advances indicate towards plausibility of deploying deep learning based Computer Aided Diagnosis (CAD) tools in real world clinical settings in the near future. However, one big challenge in this direction comes from incremental and slow annotation processes of large datasets required for training robust deep learning algorithms. Annotating medical images requires significant domain knowledge and the availability of annotated data is only possible in a gradual manner. These two factors can make developing a large enough dataset take from several months to few years. Moreover, the presence of various controlling factors e.g. addition of new data collection site, availability of domain experts make the data collection procedure quite non-uniform and the new chunks of data are available in very uncertain and dynamic manner. Given these limitations, incremental learning proves to be quite efficient solution that allows for early deployment of CAD tools and gradual improvement in performance as and when the annotated data becomes available.

The widely known challenge for incremental learning comes from subsequent forgetting as the model is tuned to adapt to new unseen data. This problem is further aggravated in medical imaging domain, where factors such as change in acquisition protocols over time, different formats of underlying data, and widely varying imbalance between classes add to the existing variance of the incoming new datasets. Therefore, simply fine-tuning the existing network on new incoming dataset would lead to gradual drift of model towards suboptimal local minima. A robust algorithm that ensures generalization of the model weights to new unseen data while retaining existing knowledge is therefore essential for availing benefits of time and resource-efficiency of the incremental learning setup.

Hard Example Mining (HEM) is widely used approach for alleviating the problem of heavy class imbalance especially observed in object detection tasks. HEM is used to select a subset of hard examples (i.e., the examples that yield high prediction error, also computed in terms of high training loss) from entire pool of examples. Training on only hard examples as compared entire dataset not only helps alleviate the problem of data imbalance, but also reduces computational cost as the network training is not inflated by easy examples that do not contribute towards increasing generalizability of the network. In this work, we develop a novel hard example mining approach to be used in incremental learning setup. We have developed and tested the proposed approach for the medical image tasks. However, the proposed approach is quite generic and can be easily utilized in other settings as well.

In summary, our approach consists of fine-tuning the model on a subset containing samples from hard as well as easy examples of each class every time a new chunk of annotated data becomes available. As we show in sections below, the proposed approach efficiently adapts to new unseen dataset in lesser time along with retaining existing knowledge. In the sections below, we first provide an overview of related work in the domain of incremental learning and hard example mining followed by description of our approach. The results section demonstrates the promising performance of our proposed approach in one real world scenario consisting of our in-house dataset and one publicly available dataset.

\section{Related Work}

\subsection{Incremental Learning}

Incremental learning is very intuitive direction towards developing next level artificial general purpose intelligence as humans and other animals can effectively learn from new data without forgetting old information. There exists considerable amount of prior work on incremental learning on deep CNNs. A large section of which belongs to learning new classes from fewer samples (\cite{Fei-Fei2006One-shotCategories}, \cite{Lampert2009LearningTransfer}) utilizing transfer learning techniques. \cite{Sarwar2017IncrementalSharing} proposed a method to incrementally grow network to learn new class while sharing part of the base network. Learning without forgetting \cite{LiLearningForgetting} is another method that uses only new data to train the network while preserving the original capabilities. In this work, the original network is trained on an extensive dataset, such as ImageNet, and the new task data is a much smaller dataset. \cite{Kochurov2018BayesianNetworks} applied incremental learning on new data using bayesian techniques. However, there is extreme scarcity of existing work in the direction of incremental learning in medical imaging domain despite the clear need for developing incremental learning capabilities in this domain.

\subsection{Hard Example Mining}
Hard example mining techniques have been widely applied to training classic models (\cite{Suykens1998LeastClassifiers}, \cite{Dollar2009IntegralFeatures}).  Boosted decision tree in \cite{Dollar2009IntegralFeatures} is trained with hard example mining strategy but hard examples are mined only once. Another hard example mining technique named bootstrapping is used to train Support Vector Machines (SVMs) \cite{Suykens1998LeastClassifiers} Further, hard example mining is used to avoid inflation of easy negative samples while training object detectors. \cite{Shrivastava2016TrainingMining} proposed an online hard example mining approach to automatically select hard examples from dataset. Loss rank mining \cite{Yu2018LossDetectors} filtered out easy examples in final feature map and forced to concentrate on hard examples during training. To the best of our knowledge, this work is first effort to apply hard example mining in the incremental learning scenario as well as in medical imaging domain.


\section{Dataset}
\subsection{Bitewing radiographs}
\label{bwxdataset}
We obtained over 6000 bitewing radiographs from approximately 150 clinics across the USA after approval from IRB. All the radiographs were annotated by certified dentists after clinical verification for the existence of dental caries. A baseline model was trained on previously available subset of the dataset \cite{Srivastava2017DetectionLearning} and under test across many clinics in USA. Further data collection and annotation was done in several stages in variable chunk sizes. Total training dataset consisted of 5000 bitewing radiographs at the final stage of incremental training. Testing dataset consisted of 1000 radiograph images separated at the initial stage of data collection.

\subsection{ISIC skin lesion}
The International Skin Imaging Collaboration (ISIC) is an international effort to improve melanoma diagnosis and provides 2000 dermoscopic images for training and 600 for testing annotated by recognized skin cancer experts. These annotations include dermoscopic features (i.e., global and focal morphologic elements in the image known to discriminate between types of skin lesions). In order to adapt the dataset for incremental learning set up, we split 2000 training images into five parts: 1 set of 1000 images and four sets of 250 images each.

\section{Approach}
We developed Hard Example Mining (HEM) inspired approach for incremental learning called Incremental Example Mining (IEM) of one of the clinically under test diagnosis tools [\href{https://dentistry.ai/}{dentistry.AI}]. Since the new data is collected from multiple dentists and from multiple machines; the annotated data can only be available chunks of variable size with varying class ratios, inter-class and intra-class variances. Therefore, the incremental learning paradigm should be designed so as to take into account all these factors while updating the model weights. A naive method of simply fine-tuning the model with new data can lead to significant model drift towards suboptimal local minima. In order to factor in all these challenges of incremental learning in already under test deep learning algorithms in clinical settings; we have designed Hard Example Mining based approach to gradually update the model weights without risking loss of its original capabilities.

Let $N_0$ be the number of initial examples of the dataset that are used to train baseline model before deployment. Let Ni be the number of examples available at $i^{th}$ stage of incremental training. For any $i^{th}$ stage, we draw a subset of examples (4*K) from entire dataset (i.e., $N_0$ + $N_1$ + $N_2$ + ... + $N_i$) to fine-tune the model at $i^{th}$ stage; where K is called Partition Number. The model at $i^{th}$ step of incremental training is fine-tuned with a set of 4*K examples drawn from entire dataset containing K hard positives, K hard negatives, K easy positives and K easy negatives. The optimal value of the hyperparameter K is determined empirically for each dataset while considering the trade-off between obtained accuracy of the fine-tuned model and computation time. The rationale behind fine-tuning on a balanced set of hard positives as well as hard negatives is to avoid disruption of model gradients from outlier hard examples.

In order to reduce the effect of outlier hard examples (i.e., hard examples that are either very different from entire dataset or contain label noise) to drift the model towards suboptimal local minima; we define an hyperparameter, dropping number (d). We maintain the record of the number of stages (C) any example was contained in the subset used to fine-tune the model. If the number exceeds dropping number, it is marked as outlier and its error term is set to zero so as to exclude it from further stages of model fine-tuning.

\begin{figure}[t]
\begin{center}
\includegraphics[width=0.9\linewidth]{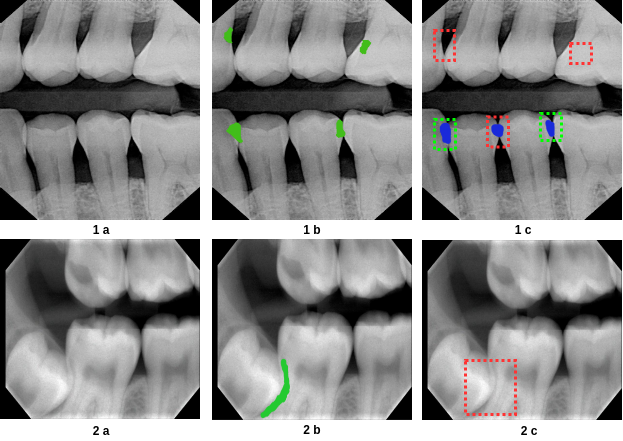}
\end{center}
   \caption{1a,2a) Raw bitewing image 1b,2b) Ground truth (green color) on the bitewing image 1c,2c) Prediction (blue color) during training on the bitewing image. Green box corresponds to correct prediction and red box correspond to incorrect and missed prediction. In both the images, the area of incorrect and missed cavities (or, loss value) are almost same but the error term gives more weightage to the fist image. }
\label{fig:example}
\end{figure}

Further, we define a robust criterion to define hard examples. We speculate that use of mean cross entropy loss across all image pixels as measure for estimating difficulty level of an example is very naive. Such a method is not able to distinguish between type I and type II errors i.e., errors caused by False Positives (FP), False negative (FN) [Figure ~\ref{fig:example}]; which is very critical for a clinical scenario where reducing one type of error may be more important than the other. In order to make our approach sensitive to all types of relevant errors, we define a composite “error term” to compute the difficulty level of each example. The “error term” is defined as:-

\begin{equation}
  E = L + FP + FN + (1 - JI)
  \label{equ:dt}
\end{equation}

Where, L is mean cross-entropy loss across all pixels in the image; FP is number of False Positives and FN is number of False negatives (computed at cavity level); JI is Jaccard Index or Intersection over Union, which can also be thought of precision at pixel level.
We update the E value of images who undergo training after every iteration. We take loss value and jaccard index to calculate E for the ISIC skin lesion dataset. 

Step by step procedure for incremental learning using IEM is explained below:

\begin{algorithm}[H]
\caption{Incremental Example Mining (IEM)}
\label{ihem}
\begin{algorithmic}[1]
\State Let M be the model
\State Let C be the count term, E be the error term, K be the partition number, d be the dropping number and t be the number of augmentation
\For{each incremental step}
    \State Initialize M with the previously saved weight
    \State Let $U_n$ be new data and $U_o$ be old data
    \State $U \leftarrow U_n + U_o$
    \State Calculate K
    \State Forward propagation on U
    \State $P \leftarrow [U, C, E]$ 
    \For{each iteration}
    \State $P \leftarrow Sort(P, E)$
    \State $pos \leftarrow U.where(label == 1)$
    \State $neg \leftarrow U.where(label == 0)$
    \If{$\textit{len(pos)} \neq \textit{len(neg)}$}
    \State $neg \leftarrow neg[0:len(pos)]$
    \EndIf
    \State $X \leftarrow [ ]$
    \State $X.append(pos[0:K])$ ; $X.append(neg[0:K])$
    \State $pos \leftarrow pos[K:len(pos)]$
    \State $neg \leftarrow neg[K:len(neg)]$
    \State $shuffle(pos$) ; $shuffle(neg)$
    \State $X.append(pos[0:K])$ ; $X.append(neg[0:K])$
    \State Start training on X
    \For{$i$ in (1, len(U))}
    \If{$U_i$ in X}
    \For{$j$ in (1, t)}
    \State Update $E_{ij}$ using Equation \ref{equ:dt}
    \State $E_i \leftarrow E_i + E_{ij}$
    \EndFor
    \State $E_i \leftarrow E_i / t$
    \State $C_i \leftarrow C_i + 1$
    \If{$C_i > d$}
    \State $E_i \leftarrow 0$
    \EndIf
    \EndIf
    \EndFor
    \EndFor
\EndFor
\end{algorithmic}
\end{algorithm}

\section{Implementation}
\subsection{Architecture}
We apply our incremental learning approach for fine-tuning our clinically under test model and one baseline model trained on partial dataset from ISIC skin lesion challenge. We choose U-Net\cite{RonnebergerU-Net:Segmentation} as the underlying architecture for ISIC skin lesion. U-Net is a popular fully convolutional networks (FCNs) architecture for biomedical image segmentation. The architecture consists of a downsampling path to encode an input image into feature representations and a upsampling path which expands the feature to same size of the input image. It can be trained end-to-end from very few images and outperforms the prior state of the art methods. The main contribution of U-Net compared to other FCNs is the use of skip connections from downsampling path to upsampling path to make use of low-level information while reconstructing the segmentation output.\\

We apply incremental learning to our in-house dataset \ref{bwxdataset}. IEM is used to incrementally train the system mentioned in \cite{Srivastava2017DetectionLearning}. This system uses U-Net to segment out areas of dental caries and a bounding box over these segmentation outputs is presented as the final output. These final outputs are compared with bounding box on test annotations for a threshold IoU measure to find hits/misses. Precision and recall are calculated to evaluate the system.

\subsection{Experiments}
We use two other approaches to compare the performance of our incremental learning approach. First, we trained a baseline network from scratch using entire dataset at once. Next, we train the network with entire dataset using hard example mining (baseline + HEM), where we followed same sampling approach (hard and easy examples) from IEM. As the empirical evidences from literature indicate, the performance of “baseline + HEM” model  should be better than the baseline method as it handles the problem of imbalance robustly along with better generalization to rare examples. We hypothesize that the incremental learning approach should perform better than the baseline model. Further, the performance of incremental learning approach should also be similar to the “baseline + HEM” model so as to be considered as a viable option in real-world.

\subsection{Training and Hyperparameter Search}
We performed random search to find optimal partition number (K) and dropping number (d). We found optimal K to be the number of positive examples in a new set of data. For example, in the ISIC skin lesion, let's call the first 1000 batch of images as “initial data” and subsequent 250 batch of images as “new data”. We train the U-Net for first 1000 images and apply IEM for succedent incremental batches of 250 images each. Suppose the 250 batch of new data contains say 100 positive examples, then the final dataset for IEM training contains 400 examples; 100 hard positives, 100 hard negatives, 100 easy positives and 100 easy negatives.
Further, the optimal d was found to be 10 iterations for both the datasets.

\section{Results}
We have tried IEM for both bitewings radiographs and ISIC skin lesion dataset. We get similar or better accuracy for both the dataset from our baseline models after applying IEM.  We have shown three results, first is our baseline models trained on entire dataset, second applying IEM on top of the baseline model for the incremental dataset and third applying HEM on top of the baseline for the entire dataset.

\subsection{Bitewing radiographs}
\setlength{\tabcolsep}{4pt}
\begin{table}[H]
\begin{center}
\caption{Bitewing radiographs}
\label{table:result1}
\begin{tabular}{llll}
\hline\noalign{\smallskip}
Method & Recall & Precision & F1 Score\\
\noalign{\smallskip}
\hline
\noalign{\smallskip}
Model & 70 & 53 & 60.32\\
Model + IEM (incremental dataset) & 73 & 53 & 61.42\\
Model + HEM (entire dataset) & 69 & 46 & 55.20\\
\hline
\end{tabular}
\end{center}
\end{table}
\setlength{\tabcolsep}{1.4pt}

\subsection{ISIC skin lesion}
\setlength{\tabcolsep}{4pt}
\begin{table}[H]
\begin{center}
\caption{ISIC skin lesion}
\label{table:result2}
\begin{tabular}{ll}
\hline\noalign{\smallskip}
Method & Jaccard Index\\
\noalign{\smallskip}
\hline
\noalign{\smallskip}
U-Net & 72.01\\
U-Net + IEM (incremental dataset) & 73.81\\
U-Net + HEM (entire dataset) & 73.86\\
\hline
\end{tabular}
\end{center}
\end{table}
\setlength{\tabcolsep}{1.4pt}

\section{Conclusion}
In this work, we combined the approach of incremental learning and hard example mining to develop a new method called Incremental Example Mining (IEM) to incrementally train any model for the incremental dataset. We get similar accuracies between IEM and model trained on the entire dataset at once. However, training the model every time from scratch when new chunks of data gets available is very cumbersome and inefficient. So, we believe IEM method would be a viable option in the real world.

{\small
\bibliographystyle{ieee}
\bibliography{references}
}

\end{document}